\documentclass[final]{cvpr}

\usepackage{times}
\usepackage{epsfig}
\usepackage{graphicx,subcaption}
\usepackage{color}
\usepackage{dirtytalk}
\usepackage{comment}
\usepackage{wrapfig}
\usepackage{caption}
\usepackage{amsmath}
\usepackage{amssymb}
\usepackage{lipsum,booktabs}

\usepackage[pagebackref=true,breaklinks=true,colorlinks,bookmarks=false]{hyperref}

\newcommand{\todo}[1]{}
\renewcommand{\todo}[1]{{\color{red} TODO: {#1}}}
\newcommand{\LMR}[1]{{$MR^{-2}$}}

\newcommand{\caltech}[1]{{Caltech~\cite{dollar2012pedestrian}}}
\newcommand{\cityperson}[1]{{CityPersons~\cite{zhang2017citypersons}}}
\newcommand{\ecp}[1]{{ECP~\cite{braun2018eurocity}}}
\newcommand{\widerperson}[1]{{Wider Pedestrian~\cite{zhang2019widerperson}}}
\newcommand{\crowdhuman}[1]{{CrowdHuman~\cite{shao2018crowdhuman}}}
\newcommand{\mobnet}[1]{{MobileNet~\cite{howard2017mobilenets}}}

\newcommand{\caltecha}[1]{{Caltech}}
\newcommand{\citypersona}[1]{{CityPersons}}
\newcommand{\ecpa}[1]{{ECP}}
\newcommand{\widerpersona}[1]{{Wider Pedestrian}}
\newcommand{\crowdhumana}[1]{{CrowdHuman}}

\usepackage{manyfoot}

\DeclareNewFootnote{A}
\DeclareNewFootnote{B}

\let\footnoteR\footnoteB
\let\footnote\footnoteA

\def\cvprPaperID{230} 
\def\confYear{CVPR 2021}

\begin{document}

\title{Generalizable Pedestrian Detection: The Elephant In The Room}

\author{Irtiza Hasan $^{1}$, Shengcai Liao$^{1, \dagger}$, Jinpeng Li$^{1}$, Saad Ullah Akram$^{2}$, Ling Shao$^{1}$\\
{\normalsize Inception Institute of Artificial Intelligence (IIAI)$^{1}$, Aalto University, Finland$^{2}$}\\
{\tt\small \{irtiza.hasan,shengcai.liao,jinpeng.li,ling.shao\}@inceptioniai.org, saad.akram@aalto.fi}}
\maketitle

\renewcommand{\thefootnote}{\arabic{footnote}}
\begin{abstract}
Pedestrian detection is used in many vision based applications ranging from video surveillance to autonomous driving. Despite achieving high performance, it is still largely unknown how well existing detectors generalize to unseen data. This is important because a practical detector should be ready to use in various scenarios in applications. To this end, we conduct a comprehensive study in this paper, using a general principle of direct cross-dataset evaluation. Through this study, we find that existing state-of-the-art pedestrian detectors, though perform quite well when trained and tested on the same dataset, generalize poorly in cross dataset evaluation. We demonstrate that there are two reasons for this trend. Firstly, their designs (e.g. anchor settings) may be biased towards popular benchmarks in the traditional single-dataset training and test pipeline, but as a result largely limit their generalization capability. Secondly, the training source is generally not dense in pedestrians and diverse in scenarios. 
Under direct cross-dataset evaluation, surprisingly, we find that a general purpose object detector, without pedestrian-tailored adaptation in design, generalizes much better compared to existing state-of-the-art pedestrian detectors. Furthermore, we illustrate that diverse and dense datasets, collected by crawling the web, serve to be an efficient source of pre-training for pedestrian detection. Accordingly, we propose a progressive training pipeline and find that it works well for autonomous-driving oriented pedestrian detection. Consequently, the study conducted in this paper suggests that more emphasis should be put on cross-dataset evaluation for the future design of generalizable pedestrian detectors. Code and models can be accessed at \url{https://github.com/hasanirtiza/Pedestron}.

\end{abstract}
\section{Introduction}
\label{sec:intro}
\footnoteR{$^{\dagger}$Corresponding author.}

\begin{figure*}[]
	\begin{center}
	    \includegraphics[width=1\linewidth]{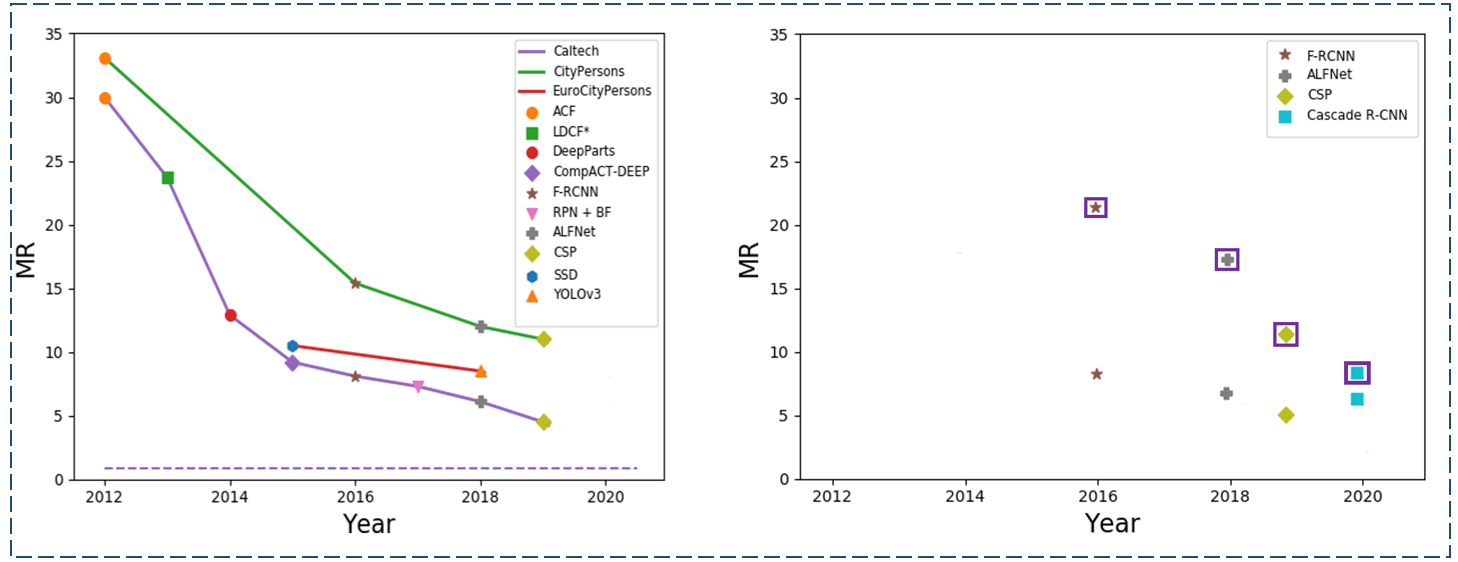}
	\end{center}
\caption{Left: 
Pedestrian detection performance over the years for \textit{Caltech}, \textit{CityPersons} and \textit{EuroCityPersons} on the reasonable subset.
\textit{EuroCityPersons} was released in 2018 but we include results of few older models on it as well.
Dotted line marks the human performance on \textit{Caltech}. Right: We show comparison between traditional single-dataset train and test evaluation on \caltech{} \emph{vs.} cross-dataset evaluation for three pedestrian detectors and one general object detector (Cascade R-CNN). Methods enclosed with bounding boxes are trained on \cityperson{} and evaluated on \caltech{}, while others are trained on \caltecha{}. }
\label{fig:progress}
\end{figure*}


Pedestrian detection is one of the longest standing problems in computer vision. Numerous real-world applications, such as, autonomous driving \cite{campmany2016gpu,hbaieb2019pedestrian}, video surveillance \cite{hattori2015learning}, action recognition \cite{zhang2020semantics} and tracking \cite{huang2019bridging} rely on accurate pedestrian/person detection. 
Recently, convolutional neural network (CNNs) based approaches have shown considerable progress in the field of pedestrian detection, where on certain benchmarks, the progress is within striking distance of a human baseline as shown in Fig. \ref{fig:progress} left.

However, some current pedestrian detection methods show signs of over-fitting to source datasets, especially in the case of autonomous driving. As shown in Fig. \ref{fig:progress} right, current pedestrian detectors, do not generalize well to other (target) pedestrian detection datasets, even when trained on a relatively large scale dataset which is reasonably closer to the target domain. This problem prevents pedestrian detection from scaling up to real-world applications. 

Despite being a key problem, generalizable pedestrian detection has not received much attention in the past. More importantly, reasons behind poor performances of pedestrian detectors in cross-dataset evaluation has not been properly investigated or discussed. In this paper, we argue that this is mainly due to the fact that the current state-of-the-art pedestrian detectors are tailored for target datasets and their overall design is biased towards target datasets, thus reducing their generalization. Secondly, the training source is generally not dense in pedestrians and diverse in scenarios. Since current state-of-the-art methods are based on deep learning, their performance depend heavily on the quantity and quality of data and there is some evidence that the performance on some computer vision tasks (e.g. image classification) keeps improving at least up-to billions of samples \cite{mahajan2018exploring}.

At present, all autonomous driving related datasets have at least three main limitations, 1) limited number of unique pedestrians, 2) low pedestrian density, i.e. the challenging occlusion samples are relatively rare, and 3) limited diversity as the datasets are captured by a small team primarily for dataset creation instead of curating them from more diverse sources (e.g. youtube, facebook, etc.).

In last couple of years, few large and diverse datasets, CrowdHuman \cite{shao2018crowdhuman}, WiderPerson~\cite{zhang2019widerpersondataset} and \widerpersona{} \cite{zhang2019widerperson}, have been collected by crawling the web and through surveillance cameras.
These datasets address the above mentioned limitations but as they are from a much broader domain, they do not sufficiently cover autonomous driving scenarios.
Nevertheless, they can still be very valuable for learning a more general and robust model of pedestrians. 
As these datasets contain more person per image, they are likely to contain more human poses, appearances and occlusion scenarios, which is beneficial for autonomous driving scenarios, provided current pedestrian detectors have the innate ability to digest large-scale data.

In this paper, we demonstrate that the existing pedestrian detection methods fare poorly compared to general object detectors when provided with larger and more diverse datasets, and that the state-of-the-art general detectors when carefully trained can significantly out-perform pedestrian-specific detection methods on pedestrian detection task, without any pedestrian-specific adaptation on the target data (see Fig. \ref{fig:progress} right).
We also propose a progressive training pipeline for better utilization of general pedestrian datasets for improving the pedestrian detection performance in case of autonomous driving.
We show that by progressively fine-tuning the models from the largest (but farthest away from the target domain) to smallest (but closest to the target domain) dataset, we can achieve large gains in performance in terms of \LMR{} on reasonable subset of Caltech (3.7$\%$) and CityPerson (1.5$\%$) without fine-tuning on target domain.
These improvement hold true for models from all pedestrian detection families that we tested such as Cascade R-CNN \cite{cai2019cascade}, Faster RCNN \cite{ren2015faster} and embedded vision based backbones such as \mobnet.

The rest of the paper is organized as follows. Section \ref{sec:prev} reviews the relevant literature. 
We introduce datasets and evaluation protocol in Sec. \ref{sec:experi}. We benchmark our baseline in Sec. \ref{sec:bench}. 
We test the generalization capabilities of the pedestrian specific and general object detectors in Sec. \ref{sec:general-cap}. 
Finally, conclude the paper in Section \ref{sec:conc}.


\section{Related Work} \label{sec:prev}

\noindent\textbf{Pedestrian detection.} Before the emergence of CNNs, a common  way to address this problem was to exhaustively operate in a sliding window manner over all possible locations and scales, inspired from Viola and Jones \cite{viola2004robust}. 
Dalal and Triggs in their landmark pedestrian detection work \cite{dalal2005histograms} proposed Histogram of Oriented Gradients (HOG) feature descriptor for representing pedestrians.
Dollar et al \cite{dollar2014fast}, proposed ACF, where the key idea was to use features across multiple channels. 
Similarly, \cite{zhang2017citypersons,paisitkriangkrai2014strengthening}, used filtered channel features and low-level visual features along with spatial pooling respectively for pedestrian detection. %
However, the use of engineered features meant very limited generalization ability and limited performance.    

In recent years, Convolutional Neural Networks (CNNs) have become the dominant paradigm in generic object detection \cite{ren2015faster,he2017mask,sun2018fishnet,liu2016ssd}.
The same trend is also true for the pedestrian detection \cite{angelova2015real,hosang2015taking,cai2015learning}. 
Some of the pioneer works for CNN based pedestrian detection \cite{hosang2015taking,zhang2016far} used R-CNN framework \cite{girshick2014rich}, which is still the most popular framework.
RPN+BF \cite{zhang2016faster} was the first work to use Region Proposal Network (RPN); it used boosted forest for improving pedestrian detection performance.
This work also pointed out some problems in the underlying classification branch of Faster RCNN \cite{ren2015faster}, namely that the resolution of the feature maps and class-imbalance. 
However, RPN+BF despite achieving good performances had a shortcoming of not being optimized end-to-end. After the initial works, Faster RCNN \cite{ren2015faster} became most popular framework with wide range of literature deploying it for pedestrian detection \cite{zhou2018bi,zhang2017citypersons,cai2016unified,brazil2017illuminating,mao2017can}. 

Some of the recent state-of-the-art pedestrian detectors include ALF \cite{liu2018learning}, CSP \cite{Liu2018DBC} and MGAN \cite{pang2019maskguided}.
ALF \cite{liu2018learning} is based on Single Shot MultiBox Detector (SSD) \cite{liu2016ssd}, it stacks together multiple predictors to learn a better detection from default anchor boxes.
MGAN \cite{pang2019maskguided} uses the segmentation mask of the visible region of a pedestrian to guide the network attention and improve performance on occluded pedestrians.
CSP is an anchor-less fully convolutional detector, which utilizes concatenated feature maps for predicting pedestrians.

\noindent\textbf{Pedestrian detection benchmarks.}
Over the years, several datasets for pedestrian detection have been created such as Town Center \cite{benfold2011stable}, USC \cite{wu2007cluster}, Daimler-DB \cite{munder2006experimental}, INRIA \cite{dalal2005histograms}, ETH \cite{ess2007depth}, and TUDBrussels \cite{wojek2009multi}. 
All of the aforementioned datasets were typically collected for surveillance application. 
None of these datasets were created with the aim of providing large-scale images for the autonomous driving systems. 
However, in the last decade several datasets have been proposed from the context of autonomous driving such as KITTI \cite{geiger2012we}, \caltech{}, \cityperson{} and \ecp{}. Typically these datasets are captured by a vehicle-mounted camera navigating through crowded scenarios. 
These datsets have been used by several methods with \caltech{} and \cityperson{} being the most established benchmarks in this domain. 
However, \caltech{} and \cityperson{} datasets are monotonous in nature and they lack diverse scenarios (contain only street view images). 
Recently, \ecp{} dataset which is an order of magnitude larger than \cityperson{} has been porposed. 
\ecp{} is much bigger and diverse as it contains images from all seasons, under both day and night times, in several different countries. 
However, despite its large scale, \ecp{} provides a limited diversity (in terms of scene and background) and density (number of people per frame is less than 10). 
Therefor, in this paper we argue that despite some recent large scale datasets, the ability of pedestrian detectors to generalize has been constrained by lack of diversity and density. 
Moreover, benchmarks such as WiderPerson~\cite{zhang2019widerpersondataset} \widerperson{} and \crowdhuman{}, which contain web crawled images provide a much larger diversity and density. 
This enables detectors to learn a more robust representation of pedestrians with increased generalization ability. 

\noindent\textbf{Cross-dataset evaluation.}
Previously, other works \cite{braun2018eurocity,shao2018crowdhuman,zhang2017citypersons} have investigated the role of diverse and dense datasets in the performance of pedestrian detectors. 
Broadly, these works focused on the aspect that how much pre-training on a large-scale dataset helps in the performance of a pedestrian detector and used cross-dataset evaluation for this task. However, in this work we adopt cross-dataset evaluation to test the generalization abilities of several state-of-the-art pedestrian detectors.
Under this principal, we illustrate that current state-of-the-art detectors lack in generalization, whereas a general object detector generalizes performs and through a progressive training pipeline significantly surpasses current pedestrian detectors. 
Moreover, we include more recent pedestrian detection benchmarks in our evaluation setup.

\section{Experiments} \label{sec:experi}
\subsection{Experimental Settings}

\begin{table*}[tb]
\centering
\caption{Datasets statistics. $\ddagger$ Fixed aspect-ratio for bounding boxes.}
\label{tab:data-stat}
\begin{tabular}{l|c| c| c| c| c}
\hline
 & Caltech $\ddagger$ & CityPersons $\ddagger$ & ECP & CrowdHuman& Wider Pedestrian\\ \hline
images &42,782  & 2,975  & 21,795 & 15,000 &90,000  \\ \hline
persons &13,674  &19,238  & 201,323 &339,565  &287,131  \\ \hline
persons/image &0.32  &6.47  & 9.2 &22.64  &3.2  \\ \hline
unique persons &1,273  &19,238  & 201,323 &339,565  & 287,131  \\ \hline
\end{tabular}
\end{table*}

\begin{table}[tb]
\centering
\caption{Experimental settings.}
\label{tab:exp-setting}
\begin{tabular}{l|c|c}
\hline
Setting             & Height & Visibility \\ \hline
Reasonable    & [50, inf]  &    [0.65, inf] \\ \hline
Small    & [50, 75] & [0.65, inf]    \\ \hline
Heavy &  [50, inf] &    [0.2, 0.65]\\ \hline
Heavy* &  [50, inf] &    [0.0, 0.65]\\ \hline
All &   [20, inf]  & [0.2, inf]  \\ \hline
\end{tabular}
\end{table}

\textbf{Datasets.} We thoroughly evaluate and compare against state-of-the-art on three large-scale pedestrian detection benchmarks. These benchmarks are recorded from the context of autonomous driving, we refer to them as \emph{autonomous driving} datasets. The \textbf{Caltech} \cite{dollar2012pedestrian} dataset has around 13K persons extracted from 10 hours of video recorded by a vehicle in Los Angeles, USA. All experiments on \caltech{} are conducted using new annotations provided by \cite{zhang2016far}.
\textbf{CityPersons} \cite{zhang2017citypersons} is a more diverse dataset compared to \caltecha{} as it is recorded in 27 different cities of Germany and neighboring countries. 
\citypersona{} dataset has roughly 31k annotated bounding boxes and its training, validation and testing sets contain 2,975, 500, 1,575 images, respectively. 
Finally, \textbf{EuroCity Persons} (ECP) \cite{braun2018eurocity} is a new pedestrian detection dataset, which surpasses Caltech and CityPersons in terms of diversity and difficulty. 
It is recorded in 31 different cities across 12 countries in Europe. It has images for both day and night-time (thus referred to as ECP day-time and ECP night-time). Total annotated bounding-boxes are over 200K. As mentioned in \ecp{}, for the sake of comparison with other approaches, all experiment and comparisons are done on the day-time \ecpa{}. We report results on the validation set of \ecp{} unless stated otherwise. Evaluation server is available for the test set and frequency submissions are limited. Finally, in our experiments we also include two non-traffic related recent datasets namely, \textbf{CrowdHuman} \cite{shao2018crowdhuman} and \textbf{Wider Pedestrian}\footnote{\widerpersona{} has images from surveillance and autonomous driving scenarios. In our experiments, we used the data provided in 2019 challenge. Data can be accessed at : \url{https://competitions.codalab.org/competitions/20132}} \cite{zhang2019widerperson}.
Collectively we refer to \caltecha{}, \citypersona{} and \ecpa{} as \emph{\textbf{autonomous driving}} datasets and \crowdhumana{}, \widerpersona{} as \emph{\textbf{web-crawled}} datasets. Details of the datasets are presented in Table \ref{tab:data-stat}.\\
\textbf{Evaluation protocol.} 
Following the widely accepted protocol of \caltech{}, \cityperson{} and \ecp{}, the detection performance is evaluated using log average miss rate over False
Positive Per Image (FPPI) over range [$10^{-2}$, $10^{0}$] denoted by (\LMR{}). 
We evaluate and compare all methods using similar evaluation settings. 
We report numbers for different occlusion levels namely, \textbf{Reasonable}, \textbf{Small}, \textbf{Heavy}, \textbf{Heavy*}\footnote{In the case of \citypersona{}, under \textbf{Heavy*} occlusion the visibility level is [0.0,0.65], for the sake of comparison with previous approaches, we used the same visibility level.} and \textbf{All} unless stated otherwise, definition of each split is given in Table \ref{tab:exp-setting}. \\
\textbf{Cross-dataset evaluation.} In cross-dataset evaluation, when written A$\rightarrow$B, we train a model only on the training set of \emph{A} and test it on the testing/validation set of \emph{B}, this training and testing routine is consistent across all experiments.   
\\
\textbf{Baseline.} Since most of the top ranked methods on \caltecha{}, \citypersona{} and \ecpa{} are direct extension of Faster/Mask R-CNN \cite{ren2015faster,he2017mask} family, we also select recent \textbf{Cascade R-CNN \cite{cai2019cascade}} (an extension of R-CNN family) as our \textbf{baseline}. 
In text, we interchangeably use baseline and Cascade RCNN, they both refer to exactly the same method Cascade R-CNN \cite{cai2019cascade}.
Cascade R-CNN contains multiple detection heads in a sequence, which progressively try to filter out harder and harder false positives. We tested several backbones with our baseline detector as shown in Table \ref{tab:backbone}. HRNet \cite{wang2019deep} and ResNeXt \cite{xie2017aggregated} are two top performing backbones.  
We choose HRNet \cite{wang2019deep} as our backbone network. Better performance of HRNet \cite{wang2019deep} can be attributed to the fact that it retains feature maps at higher resolution, reducing the likelihood of important information being lost in repeated down-sampling and up-sampling, which is especially beneficial for pedestrian detection where the most difficult samples are very small.\\

\begin{table*}[t]
\centering
\caption{Evaluating generalization abilities of different backbones using our baseline detector.}
\label{tab:backbone}
\begin{tabular}{l|c|c|c}
\hline
Backbone & Training & Testing &  Reasonable  \\ \hline
HRNet  & WiderPedestrian + CrowdHuman  &CityPersons& \textbf{12.8}   \\ \hline
ResNeXt  & WiderPedestrian + CrowdHuman  &CityPersons& \textbf{12.9}  \\ \hline
Resnet-101  & WiderPedestrian + CrowdHuman  &CityPersons& 15.8   \\ \hline
ResNet-50  & WiderPedestrian + CrowdHuman & CityPersons& 16.0    \\ \hline
\end{tabular}
\end{table*}

\section{Benchmarking}
\label{sec:bench}
\vspace{-2mm}
First, we present the benchmarking of our Cascade R-CNN \cite{cai2019cascade} on three autonomous driving datasets. 
Table \ref{tab:all-evalbench} presents benchmarking on \caltech{} dataset, \cityperson{} and on \ecp{} respectively. In the case of \caltecha{} and \citypersona{}, our baseline (Cascade R-CNN \cite{cai2019cascade}) without \say{bells and whistles} performs comparable to the existing state-of-the-art, which are tailored for pedestrian detection tasks. 
Its performance has a greater improvement compared to other methods with increasing dataset size.
Its relative performance is the worst on the smallest dataset (Caltech) and the best on the largest dataset (EuroCityPersons).

\begin{table}[tb]
\centering
 \caption{Benchmarking on autonomous driving datasets.}
  \label{tab:all-evalbench}
  \resizebox{0.99\linewidth}{!}{
\begin{tabular}{l|c|c|c|c}
\hline
Method     & Testing    & Reasonable & Small &Heavy   \\ \hline
ALFNet \cite{liu2018learning} & Caltech  & 6.1     &  7.9 & 51.0   \\ \hline
Rep Loss \cite{wang2018repulsion}&    Caltech    & \textbf{5.0}   & \textbf{5.2}    & 47.9   \\ \hline
CSP \cite{Liu2018DBC}  &Caltech & \textbf{5.0}     &  6.8  & \textbf{46.6}   \\ \hline
Cascade R-CNN \cite{cai2019cascade}    &  Caltech  & 6.2       & 7.4 & 55.3   \\ \hline

\hline \hline
RepLoss \cite{wang2018repulsion} &CityPersons &13.2 & - & - \\
\hline
ALFNet \cite{liu2018learning} & CityPersons&12.0 & 19.0 & 48.1  \\
\hline
CSP \cite{Liu2018DBC} &CityPersons& \textbf{11.0} & 16.0 &39.4 \\
\hline
Cascade R-CNN \cite{cai2019cascade} & CityPersons& {11.2} & \textbf{14.0} & \textbf{37.1} \\
\hline

\hline \hline
Faster R-CNN \cite{braun2018eurocity}   &ECP& 7.3  & 16.6     & 52.0 \\ \hline
YOLOv3 \cite{braun2018eurocity}    &ECP&  8.5 & 17.8     & 37.0\\ \hline
SSD \cite{braun2018eurocity} &  ECP&10.5 & 20.5   & 42.0 \\ \hline
Cascade R-CNN \cite{cai2019cascade} &ECP &\textbf{6.6}    & \textbf{13.6}   & \textbf{33.3} \\ \hline
\end{tabular}
}
\end{table}

\section{Generalization Capabilities} \label{sec:general-cap}
As discussed in the previous sections, traditionally, pedestrian detectors have been evaluated using the classical within-dataset evaluation, i.e., they are trained and tested on the same dataset. We find that existing methods may over-fit on a single dataset, and so we suggest to put more emphasis on cross-dataset evaluation for a new way of benchmarking. Cross-dataset evaluation is an effective way of testing how well a given method adapts to unseen domain. Therefore, in this section we evaluated the robustness of each method using cross-dataset evaluation.

\begin{table*}[t]
\centering
\caption{Cross dataset evaluation on Caltech and CityPersons. A$\rightarrow$B refers to training on \emph{A} and testing on \emph{B}.}
\label{tab:cross-cal-cp}
\resizebox{0.99\linewidth}{!}{
\begin{tabular}{l|c|c|c|c|c}
\hline
Method & Bakcbone &CityPersons$\rightarrow$CityPersons  & CityPersons$\rightarrow$Caltech & 
Caltech$\rightarrow$Caltech & Caltech$\rightarrow$CityPersons \\ \hline
FRCNN \cite{zhang2017citypersons} &VGG-16& 15.4  & 21.1 & 8.7 & 46.9  \\ \hline
Vanilla FRCNN \cite{zhang2017citypersons}  &VGG-16& 24.1  & 17.6 & 12.2 & 52.4  \\ \hline
ALFNET \cite{liu2018learning} &ResNet-50& 12.0  & 17.8& 6.1  & 47.3 \\ \hline
CSP \cite{Liu2018DBC} &ResNet-50& 11.0  & 12.1 & 5.0 & 43.7  \\ \hline
PRNet \cite{song2020progressive} &ResNet-50&  10.8 & 10.7& - & - \\ \hline
BGCNet \cite{li2020box} &HRNet & \textbf{8.8} & 10.2 & \textbf{4.1} & 41.4 \\ \hline
Faster R-CNN \cite{ren2015faster} &ResNext-101& 16.4 & 11.8 &  9.7 & 40.8   \\ \hline
Cascade R-CNN \cite{cai2019cascade} &HRNet&  11.2 & \textbf{8.8} & 6.2  & \textbf{36.5} \\ \hline

\end{tabular}
}
\end{table*}


\subsection{Cross Dataset Evaluation of Existing State-of-the-Art} \label{subsec:cross dataset}
In this section we demonstrate that existing state-of-the art pedestrian detectors generalize worse than general object detector. 
We show that this is mainly due to the biases in the design of methods for the target set, even when other factors, such as backbone, are kept consistent.  

To see how well state-of-the-art pedestrian detectors generalize to different datasets, we performed cross dataset evaluation of five state-of-the-art pedestrian detectors and our baseline (Cascade RCNN) on \cityperson{} and \caltech{} datasets. 
We evaluated recently proposed BGCNet \cite{li2020box}, CSP \cite{Liu2018DBC}, PRNet \cite{song2020progressive}, ALFNet \cite{liu2018learning} and FRCNN \cite{zhang2017citypersons}(tailored for pedestrian detection). 
Furthermore, we added along with baseline, \emph{Faster R-CNN \cite{ren2015faster}}, without \say{bells and whistles}, but with a more recent backbone ResNext-101 \cite{xie2017aggregated} with FPN \cite{lin2017feature}. Moreover, we implemented a vanilla \emph{FRCNN \cite{zhang2017citypersons}}  with VGG-16 \cite{simonyan2014very} as a backbone and with no pedestrian specific adaptations proposed in \cite{zhang2017citypersons} (namely quantized anchors, input scaling, finer feature stride, adam solver, ignore region handling, etc).

We present results for \caltecha{} and \citypersona{} in Table \ref{tab:cross-cal-cp}, respectively. 
We also report results when training is done on target dataset for readability purpose. For our results presented in Table \ref{tab:cross-cal-cp} (Fourth column, CityPersons$\rightarrow$Caltech), we trained each detector on \citypersona{} and tested on \caltecha{}. Similarly, in the last column of the Table \ref{tab:cross-cal-cp}, all detectors were trained on the \caltecha{} and evaluated on \citypersona{} benchmark. 
As expected, all methods suffer a performance drop when trained on \citypersona{} and tested on \caltecha{}. Particularly, BCGNet \cite{li2020box}, CSP \cite{Liu2018DBC}, ALFNet \cite{liu2018learning} and FRCNN \cite{zhang2017citypersons} degraded by more than 100 $\%$ (in comparison with fifth column, Caltech$\rightarrow$Caltech). 
Whereas in the case of Cascade R-CNN \cite{cai2019cascade}, performance remained comparable to the model trained and tested on target set. 
Since, \citypersona{} is a relatively diverse and dense dataset in comparison with \caltecha{}, this performance deterioration cannot be linked to dataset scale and crowd density. 
This illustrates better generalization ability of general object detectors over state-of-the-art pedestrian detectors. Moreover, it is noteworthy that BGCNet \cite{li2020box} like the Cascade R-CNN \cite{cai2019cascade}, also uses HRNet \cite{wang2019deep} as a backbone, making it directly comparably to the Cascade R-CNN \cite{cai2019cascade}.  

Importantly, pedestrian specific FRCNN \cite{zhang2017citypersons} performs worse in cross dataset (fourth column only), compared with its direct variant vanilla FRCNN. The only difference between between the two being pedestrian specific adaptations for the target set, highlighting the bias in the design of tailored pedestrian detectors. 

Similarly, standard Faster R-CNN \cite{ren2015faster}, though performs worse than FRCNN \cite{zhang2017citypersons} when trained and tested on the target dataset, it performs better than FRCNN \cite{zhang2017citypersons} when it is evaluated on \caltecha{} without any training on \caltecha{}.

It is noteworthy that Faster R-CNN \cite{ren2015faster} outperforms state-of-the-art pedestrian detectors (except for BGCNet \cite{li2020box}) as well in cross dataset evaluation, presented in Table \ref{tab:cross-cal-cp}. 
We again attribute this to the \emph{bias} present in the design of current state-of-the-art pedestrian detectors, which are tailored for specific datasets and therefore limit their generalization ability. 
Moreover, a significant performance drop for all methods (though ranking is preserved except for vanilla FRCNN), including Cascade R-CNN \cite{cai2019cascade}, can be seen in Table \ref{tab:cross-cal-cp}, last column. However, this performance drop is attributed to lack of diversity and density of the \caltecha{} dataset. \caltecha{} dataset has less annotations than \citypersona{} and number of people per frame is less than 1 as reported in Table \ref{tab:data-stat}. However, still it is important to highlight, even when trained on a limited dataset, usually general object detectors are better at generalization than state-of-the-art pedestrian detectors. Interestingly, Faster R-CNN's \cite{ren2015faster} error is nearly twice as high as that of BGCNet \cite{li2020box} in within-dataset evaluation, whereas it outperforms in BGCNet \cite{li2020box} in cross-dataset evaluation.

As discussed previously, most pedestrian detection methods are extensions of general object detectors (FR-CNN, SSD, etc.). However, they adapt to the task of pedestrian detection. 
These adaptations are often too specific to the dataset or detector/backbones (e.g. anchor settings \cite{zhang2017citypersons, liu2018learning}, finer stride \cite{zhang2017citypersons}, additional annotations \cite{zhou2018bi,pang2019maskguided}, constraining aspect-ratios and fixed body-line annotation \cite{Liu2018DBC,li2020box} etc.). These adaptations usually limit the generalization as shown in Table \ref{tab:cross-cal-cp}, also discussed, task specific configurations of anchors limits generalization as discussed in \cite{liu2019center}.

\subsection{Autonomous Driving Datasets for Generalization} \label{gen:ad}

\begin{table*}[tb]
\centering
\caption{Cross dataset evaluation of (Casc. R-CNN and CSP) on Autonomous driving benchmarks. Both detectors are trained with HRNet as a backbone.}
\label{tab:cross-cp-ecp-cal}
\begin{tabular}{l|c|c|c|c|c}
\hline
Method & Training   & Testing          & Reasonable & Small & Heavy \\ \hline
Casc. RCNN & CityPersons & CityPersons  & {11.2}  & 14.0     & {37.0} \\ \hline
CSP & CityPersons & CityPersons  & \textbf{9.4}  & \textbf{11.4}     & \textbf{36.7} \\ \hline
Casc. RCNN  & ECP   & CityPersons&  {10.9} & \textbf{11.4}     & 40.9\\ \hline
CSP & ECP   & CityPersons&  {11.5} & {16.6}     & 38.2\\ \hline
\hline 
Casc. RCNN  & ECP    &ECP& \textbf{6.9}  & \textbf{12.6}     & \textbf{33.1} \\ \hline
CSP & ECP    &ECP& {19.4}  & {50.4}     & {57.3} \\ \hline
Casc. RCNN & CityPersons & ECP   &  17.4 & 40.5     & 49.3\\ \hline
CSP & CityPersons & ECP   &  19.6 & 51.0     & 56.4\\ \hline
\hline 
Casc. RCNN & CityPersons & Caltech  & 8.8    &  9.8    & \textbf{28.8} \\ \hline
CSP & CityPersons & Caltech  & 10.1    &  13.3    & {34.4} \\ \hline
Casc. RCNN & ECP & Caltech& \textbf{8.1}   & \textbf{9.6}   & 29.9 \\ \hline
CSP & ECP & Caltech&10.4   & 13.7   & 31.3 \\ \hline
\end{tabular}
\end{table*}


We illustrate that even when training dataset is as large as \ecpa{} and testing set is as small as \caltecha{}, general object detection methods are better at learning a generic representation for pedestrians compared to existing pedestrian detectors (such as CSP\cite{Liu2018DBC}). Moreover, large scale dense autonomous driving datasets provide better generalization abilities.  

As illustrated in Section \ref{subsec:cross dataset}, cross dataset evaluation provides insights on the generalization abilities of different methods. However, another vital factor in generalization is dataset itself. A diverse dataset should capture the true essence of real world without bias \cite{braun2018eurocity}, detector trained on such dataset should be able to learn a generic representation that should handle subtle shifts in domain robustly. 
Deviating from previous studies \cite{braun2018eurocity,shao2018crowdhuman,zhang2019widerpersondataset} on the role of dataset in generalization, we perform a line by line comparison between state-of-the-art pedestrian detector and a general object detector when trained and tested on different datasets. In order to provide level playing field, we replace ResNet-50 in CSP \cite{Liu2018DBC} with a more powerful and recent backbone HRNet \cite{wang2019deep}. HRNet's effectiveness can be observed in Table \ref{tab:cross-cp-ecp-cal}, second row, where an improvement of 1.6$\%$ (11.0 \emph{vs.} 9.4 ) in \LMR can be seen. 


We begin by using the largest dataset in terms of diversity (more countries and cities included) and pedestrian density from the context of autonomous driving, \ecpa{}, for training and evaluate both Cascade RCNN and CSP on \citypersona{} (Table \ref{tab:cross-cp-ecp-cal} third and fourth row respectively). It can be seen that Cascade RCNN adapts better on \citypersona{}, compared to CSP (Reasonable setting), provided the same backbone. \ecpa{} is large scale dataset and intuitively one would expect CSP to outperform Cascade RCNN, since in within-dataset evaluation, CSP is better by significant margin (nearly 2$\%$ \LMR ~~points).

Furthermore, we swapped our training and testing set, and evaluated on \ecp{}. Cascade RCNN adapted better than CSP, even when the training source is not diverse. Besides \emph{Reasonable} setting, the difference between the performances are at least 5 $\%$ \LMR ~~points (across small scale pedestrians,  its 10.5 $\%$ \LMR~). Lastly, we fixed the smallest dataset \caltecha{} as our testing set and used both \ecpa{} and \citypersona{} as our training source. Last four rows of Table \ref{tab:cross-cp-ecp-cal}, illustrates the robustness of a Cascade RCNN across all settings. Importantly, when trained on a dense and diverse dataset \ecpa, Cascade RCNN has more ability to learn a better representation than CSP across all settings.

\begin{table*}[tb]
\centering
\caption{Benchmarking with CrowdHuman and Wider Pedestrian dataset.}
\label{tab:icon-ch-wider}
\begin{tabular}{l|c|c|c|c|c}
\hline
Method & Training & Testing             & Reasonable & Small & Heavy \\ \hline
Casc. RCNN & \crowdhumana{}& \caltecha{}    & 3.4  &   11.2   & 32.3 \\ \hline
CSP & \crowdhumana{}& \caltecha{}    &  4.8 &  5.7    & 31.9 \\ \hline
Casc. RCNN& \crowdhumana{}& \citypersona{}    & 15.1   & 21.4      & 49.8 \\ \hline
CSP & \crowdhumana{}& \citypersona{}    & 11.8   & 18.3      & 44.8 \\ \hline
Casc. RCNN & \crowdhumana{}& \ecpa{} & 17.9  & 36.5   & 56.9  \\ \hline

CSP & \crowdhumana{}& \ecpa{} & 19.8  & 48.9   & 60.1  \\ \hline
\hline
Casc. RCNN & \widerpersona{}&\caltecha{}    & 3.2  &   10.8   & 31.7 \\ \hline
CSP & \widerpersona{}&\caltecha{}    & 3.4  &    3.0  & 29.5 \\ \hline
Casc. RCNN& \widerpersona{}&\citypersona{}    & 16.0   & 21.6      & 57.4 \\ \hline
CSP & \widerpersona{}&\citypersona{}    & 17.0   & 22.4 & 58.2 \\ \hline
Casc. RCNN & \widerpersona{}&\ecpa{} & 16.1  & 32.8   & 58.0  \\ \hline
CSP & \widerpersona{}&\ecpa{} &  24.1 & 62.6   & 76.7  \\ \hline
\end{tabular}
\end{table*}

\begin{table*}[tb]
\centering
\caption{Investigating the effect on performance when CrowdHuman, Wider Pedestrian and ECP are merged and Cascade R-CNN \cite{cai2019cascade} is trained only on the merged dataset.}
\label{tab:icon-collapsed}
\begin{tabular}{l|c|c|c|c|c}
\hline
Method & Training        & Testing     & Reasonable & Small & Heavy \\ \hline
Casc. RCNN&CrowdHuman $\rightarrow$ ECP & CP & {10.3}     & {12.6}   & {40.7}  \\ \hline
Casc. RCNN&Wider Pedestrian $\rightarrow$ ECP & CP & \textbf{9.7}     & \textbf{11.8}   & \textbf{37.7}  \\ \hline
Casc. RCNN&Wider Pedestrian + CrowdHuman + ECP & CP & 10.9     & 12.7   & 43.1  \\ \hline
Casc. RCNN&Wider Pedestrian + CrowdHuman $\rightarrow$ ECP & CP & {9.7}     & {12.1}   & {39.8}  \\ \hline
\hline \hline 
Casc. RCNN&CrowdHuman $\rightarrow$ ECP & Caltech & {2.9}     & {11.4}   & {30.8}  \\ \hline
Casc. RCNN &Wider Pedestrian $\rightarrow$ ECP & Caltech & \textbf{2.5}     & \textbf{9.9}   & \textbf{31.0}  \\ \hline
\end{tabular}
\end{table*}

\subsection{Diverse General Person Detection Datasets for Generalization} \label{gen:div}
In this section, we investigated how well diverse and dense datasets improve generalization. We conclude, In the case of small autonomous driving datasets, such as \caltech{}, training on diverse and dense sources, which may be further away from the target domain can also benefit. However, in the case of large scale target sets, training on sources close to target domains are more effective. General object detection methods, such as cascade RCNN tend to benefit more from diverse and dense datasets than a pedestrian detector such as CSP. 

Table \ref{tab:icon-ch-wider}, presents results of pre-training of Cascade R-CNN \cite{cai2019cascade} and CSP \cite{Liu2018DBC} (HRNet \cite{wang2019deep} as a backbone) on \crowdhuman{} and \widerperson{} datasets.  
These two datasets are different from autonomous driving datasets, as \crowdhuman{} contain web-crawled images of persons in different scenarios and \widerperson{} contains images from surveillance cameras and street view images (not just street view images, making them both diverse and dense).
Since the autonomous driving datasets (\caltech{},~\cityperson{} and \ecp{}) lack in density and diversity \cite{zhang2019widerperson}, \crowdhuman{} and \widerperson{} are a suitable choice for pre-training, since average person per frame and crowd density is much larger in \crowdhuman{} and \widerperson{} combines street view images and surveillance cameras based images, adding a different form of diversity. 
In Table \ref{tab:icon-ch-wider}, it can be observed that training on \crowdhuman{} and \widerperson{} can reduce nearly half of the error on \caltecha{}{} dataset for Cascade RCNN, outperforming previous state-of-the-art, that are trained only on \caltecha{}{}. Performance improvement is also consistent in CSP \cite{Liu2018DBC}, though the margin of improvement is less than that of a general object detector. On \cityperson{}, training on \crowdhuman{} does not improve the performance for CSP \cite{Liu2018DBC} or Cascade RCNN, since, \cityperson{} is a relatively challenging dataset compared to \caltech{} (in terms of density and diversity), and requires training on sources closer to the domain. This trend can also be seen in the case of \ecp{}, where for Cascade RCNN and CSP \cite{Liu2018DBC}, the performance is lower when trained on \crowdhuman{}, compared to training on \cityperson{} as in Table \ref{tab:cross-cp-ecp-cal}. Interestingly, in the case of \widerperson{} (bottom half Table \ref{tab:icon-ch-wider}), besides \cityperson{}, the relative improvements in the case of \widerperson{} is relatively larger for general object detector. The potential reason is that compared with \crowdhuman{}, \widerperson{} is large scale and closer to the target domain. Since it contains images essentially for two views (street view and surveillance), where as \crowdhuman{} contains web-crawled person images appearing in different poses and scenes.

\subsection{Progressive Training Pipeline}

We conducted experiments to show that performance can be significantly improved through progressive fine-tuning, where starting from a general diverse dataset (farther from target domain), and subsequently fine-tuning on dataset closer to the target domain.  

With the study presented in above sections, we conduct additional experiments on the importance of progressive training. To be consistent, we do not fine-tune on the target set and for training we only use the training subset of each respective dataset. \emph{A} $\rightarrow$ \emph{B} refers to pre-training on dataset \emph{A} and fine-tuning on \emph{B}.
Whereas, \emph{A} + \emph{B} refers to simply merging the two datasets together and training the model on merged larger set. For our results presented in Table \ref{tab:icon-collapsed}, we used \cityperson{} and \caltech{} as our testing sets. It can be seen, in Table \ref{tab:icon-collapsed}, first two rows, that progressive training pipeline significantly improves the performance of Cascade RCNN. Particularly, pre-training on \widerperson{} and fine-tuning on \ecp{} brings Cascade RCNN on par with other state-of-the-art approaches on \cityperson{}, without training on the \cityperson{}. Similarly, in the case of \caltech{} as well, progressive training, outperformed previously established state-of-the-art on \caltech{} dataset. Noteworthy is the fact that performance on \caltech{} is within a close vicinity of a human-baseline (0.88).   

Finally, concatenating all datasets (Table \ref{tab:icon-collapsed}, third and fourth row), leads to improvement in performance, but it is still slightly worse than the progressive training that we have used, where we fine-tune on the autonomous driving benchmark.
The results illustrate that this strategy enables us to significantly improve the performances of state-of-the-art without fine-tuning on the actual target set. This illustrates the generalization capability of the proposed approaches can be enhanced by progressive training strategy, without exposure to the target set, Cascade R-CNN \cite{cai2019cascade} is on par with top performer on \citypersona{} and best performing on \caltech{}.

\subsection{Application Oriented Models}

For real-world applications, we show that even with a light-weight backbone architecture, by pre-training on diverse and dense datasets, \mobnet{} performs competitively to state-of-the-art method on \cityperson{}, i.e. CSP \cite{Liu2018DBC}. 

In many pedestrian detection applications, such as autonomous driving and cameras mounted on drones to localize persons, the size and computational cost of models is constrained.
We experiment with a small and light-weight model MobileNet \cite{howard2017mobilenets} v2, which is designed for mobile and embedded vision applications, to investigate if with progressive training pipeline, with a light backbone the performance improvements hold true.

Table \ref{tab:mobileNet} shows results on \cityperson{} using \mobnet{} as a backbone network architecture into Cascade R-CNN \cite{cai2019cascade}. First row of Table \ref{tab:mobileNet}, is for reference when, \mobnet{} trained and evaluated on \cityperson{}. 
Intuitively, \mobnet{} performs worse than the HRNet \cite{wang2019deep}. 
However, in the case of \mobnet{} as well, we see pre-training on \crowdhuman{} and fine-tuning on \ecp{} improves the performance of the \mobnet{}. 
Furthermore, we replaced \crowdhuman{} with \widerperson{} as the initial source of pre-training. Improvement over the Cascade R-CNN \cite{cai2019cascade} (1st row) can be observed (3rd row), where with \widerperson{} and fine-tuning on \ecp{} a performance gain of 0.6$\%$ \LMR can be seen. 
This is consistent with our previous finding reported in Table \ref{tab:icon-ch-wider}, \widerperson{} is a better source of pre-training than \crowdhuman{}, since it has images of autonomous driving scenes as well making it more closer to the target domain than \crowdhuman{}. 
Interestingly, in the case of \crowdhuman{} and \widerperson{}, even with a light-weight architecture, Cascade R-CNN \cite{cai2019cascade} is comparable state-of-the-art pedestrian detector CSP \cite{Liu2018DBC} (ResNet-50).

\begin{table}[tb]
\centering
\caption{Investigating  the performance of embedded vision model, when pre-trained on diverse and dense datasets.}
\label{tab:mobileNet}
 \resizebox{0.99\linewidth}{!}{
\begin{tabular}{l|c|c|c|c}
\hline
Training     & Testing        & Reasonable & Small & Heavy \\ \hline
CP   & CP & 12.0     & 15.3     & 47.8 \\ \hline \hline
ECP&  CP &    19.1 & 19.3     &  51.3\\ \hline
CrowdHuman$\rightarrow$ECP&  CP &    11.9 & 15.7     &  48.9\\ \hline
Wider Pedestrian$\rightarrow$ECP  & CP & \textbf{11.4}    & \textbf{14.6}    & \textbf{43.4} \\ \hline
\end{tabular}
}
\end{table}

\section{Conclusions} \label{sec:conc}

Encouraged by the recent progress of pedestrian detectors on existing benchmarks from the context of autonomous driving, we assessed real-world performance of several state-of-the-art pedestrian detectors using standard cross-dataset evaluation. We came to the conclusion that current state-of-the-art pedestrian detectors, despite achieving impressive performances on several benchmarks, poorly handle even small domain shifts.
This is due to the fact that the current state-of-the-art pedestrian detectors are tailored for target datasets and their overall design contains biasness towards target datasets, thus reducing their generalization.
In contrast, general object detectors are more robust and generalize better to new datasets. We thoroughly investigated and verified that general object detectors due to generic design can benefit more from large-scale datasets diverse in scenes and dense in pedestrians. Besides, a progressive training  pipeline proposed in this paper works well for autonomous-driving  oriented  pedestrian  detection. In summary, our findings in this paper can serve as a stepping stone in developing new generalizable pedestrian detectors. 


{\small
\bibliographystyle{ieee_fullname}
\bibliography{main}
}

\end{document}